\documentclass[11pt,reqno,a4paper]{amsart}
\usepackage{enumerate}
\usepackage{graphicx}
\usepackage[utf8]{inputenc}
\usepackage{amsmath,amssymb,MnSymbol,dsfont,mathrsfs}
\usepackage{hyperref}
\usepackage{doi}
\usepackage[numbers,sort&compress]{natbib}
\usepackage{multicol}
\usepackage{tgtermes}
\usepackage[table,x11names]{xcolor}

\usepackage[margin=2cm]{geometry}

\newcommand{\eps}{\varepsilon}
\newcommand{\EE}{\mathbb{E}}
\newcommand{\RR}{\mathds{R}}

\begin{document}

\title{Sampling conditioned diffusions via Pathspace Projected Monte Carlo}
\let\thefootnote\relax\footnotetext{Tobias Grafke \textit{Date:}
  \today} 

\begin{abstract}
  We present an algorithm to sample stochastic differential equations
  conditioned on rather general constraints, including integral
  constraints, endpoint constraints, and stochastic integral
  constraints. The algorithm is a pathspace Metropolis-adjusted
  manifold sampling scheme, which samples stochastic paths on the
  submanifold of realizations that adhere to the conditioning
  constraint. We demonstrate the effectiveness of the algorithm by
  sampling a dynamical condensation phase transition, conditioning a
  random walk on a fixed Levy stochastic area, conditioning a
  stochastic nonlinear wave equation on high amplitude waves, and
  sampling a stochastic partial differential equation model of
  turbulent pipe flow conditioned on relaminarization events.
\end{abstract}

\maketitle

\section{Background}
\label{sec:background}

Stochastic differential equations form the basis of a multitude of
models in the sciences: The degrees of freedom under consideration
evolve according to a known and deterministic model, and all
uncertainty or unresolved processes can be added as stochasticity. In
many cases of interest, one is then interested in the probability
distribution of observables of the model conditional on some
physically meaningful event. Examples include teleconnection patterns
during heat waves~\cite{ragone-wouters-bouchet:2018,
  ragone-bouchet:2021}, the pathway to the collapse of the Atlantic
meridional overturning
circulation~\cite{lohmann-dijkstra-jochum-etal:2024,
  soons-grafke-dijkstra:2024} or other climate tipping
points~\cite{ashwin-wieczorek-vitolo-etal:2012}, the probability of
large ``rogue'' waves in the
ocean~\cite{onorato-residori-bortolozzo-etal:2013,
  dematteis-grafke-onorato-etal:2019}, or the expected waiting time
for rare events in fluids, such as the proliferation of turbulence in
pipe flow~\cite{gome-tuckerman-barkley:2022, frishman-grafke:2022-a},
or the rupture of a liquid
nanofilm~\cite{zhang-sprittles-lockerby:2019,
  sprittles-liu-lockerby-etal:2023, liu-sprittles-grafke:2024}. In all
these cases, we are interested in drawing realizations of the
stochastic process, but conditioned on a particular outcome. This
problem is most difficult if the particular outcome is \emph{rare} (as
in all examples above), such that naively sampling the process many
times will yield almost no realizations of the constraint, and thus
any conditional statistics are very hard to obtain.

The answer to this problem are specifically crafted \emph{rare event
algorithms}. This contains methods to directly bias the dynamics to
observe certain rare events (which falls under the class of importance
sampling Monte Carlo algorithms), or alternatively \emph{effectively}
bias an ensemble of simulations by subselection, cloning, and pruning
(also known as importance splitting, such as sequential Monte-Carlo,
multilevel splitting~\cite{brehier-lelievre-rousset:2015}, or
genealogical particle
algorithms~\cite{giardina-kurchan-lecomte-etal:2011}). Opposed to this
are sampling free rare events methods, for example based on large
deviation theory, to estimate rare events by their most likely
singular occurrence~\cite{grafke-vanden-eijnden:2019}. Lastly, one can
directly draw from the dynamics conditioned on the rare outcome, which
includes theoretical tools like the Doob
transform~\cite{chetrite-touchette:2015}, as well as numerical methods
such as transition path
sampling~\cite{bolhuis-chandler-dellago-etal:2002,
  bolhuis-swenson:2021, grafke-laio:2024} and bridge
sampling~\cite{beskos-roberts-stuart-etal:2008,
  cotter-roberts-stuart-etal:2013, bierkens-meulen-schauer:2020,
  grong-habermann-sommer:2024}. Here, we will consider an algorithm of
the latter class, but not only for bridges, or transitions from one
point in the state space to another, but on rather arbitrary
observables, including time-integrated ones, those that are the result
of stochastic integrals along the process, or observables that are a
nonlinear function of the final realized configuration after a random
temporal evolution.

The algorithm falls into the class of Metropolis-adjusted pathspace
Monte Carlo samplers~\cite{rossky-doll-friedman:1978,
  roberts-tweedie:1996, stuart-voss-wilberg:2004,
  hairer-stuart-voss:2007, beskos-roberts-stuart-etal:2008}, but
realized on manifolds, where the manifold is the set of random
parameters that realize the constraint. While we restrict it here to
additive and multiplicative Gaussian stochastic (partial) differential
equations, the generalization to arbitrary stochastic processes with
random parameters is straightforward.

For this, we will introduce our considered setup in
section~\ref{sec:setup}. Then, in section~\ref{sec:pathsp-metr-hast}
we will present our algorithm, by first discussing existing pathspace
sampling and diffusion bridge sampling ideas in
section~\ref{sec:pathsp-lang-mcmc}, and then developing our algorithm
in sections~\ref{sec:cond-sampl-as}
to~\ref{sec:backpr-onto-mathc}. Following that, in
section~\ref{sec:examples}, we will demonstrate the applicability of
the algorithm to various problems, namely range statistics of the
standard Brownian bridge in section~\ref{sec:range-stat-stand}, a
dynamical condensation phase transition in
section~\ref{sec:dynam-cond-phase}, the stochastic Levy area in
section~\ref{sec:stochastic-levy-area}, a PDE case of large waves in
the nonlinear Korteweg-deVries wave equation in
section~\ref{sec:pde-with-endpoint}, and lastly the decay of turbulent
puffs in a model of subcritical pipe flow in
section~\ref{sec:decay-turb-puffs}. We close off with a discussion in
section~\ref{sec:discussion}. Some issues and open questions for the
scalability of the algorithm in infinite dimensional Wiener space is
discussed in appendix~\ref{sec:metr-hast-wien}.

\section{Setup}
\label{sec:setup}

Consider the (additive Gaussian) SDE for $X_t$ on $\RR^n$ with
$t\in[0,T]$ given by
\begin{equation}
  \label{eq:SDE}
  dX_t = b(X_t)\,dt + \sigma dW_t\,,\qquad X_0=x_0\,,
\end{equation}
for some drift vector field $b : \RR^n\to\RR^n$, and diffusion matrix
$\sigma \in \RR^{n\times m}$, with $W_t$ being $m$-dimensional
Brownian motion. For notational purposes, we define the ordinary
differential equation
\begin{equation}
  \label{eq:phi-map}
  \dot\phi(t) = b(\phi) + \sigma\eta
\end{equation}
for $\phi:[0,T]\to\RR^n$ that maps some noise realization $\eta\in
L^2([0,T],\RR^m)$ to an outcome trajectory, and associate it with the
solution map $\Phi : \eta \mapsto \phi$. Then, we define as $f(\phi):
L^2([0,T],\RR^n)\to\RR$ the \emph{observable} that maps a trajectory to our desired
outcome (for example measuring properties of the final configuration
$\phi(T)$, or some integral constraint of $\int_0^T g(\phi(t))\,dt$).
Similarly, the \emph{noise-to-event} map $F = f\circ \Phi$ maps a
noise realization directly to the outcome, through the implicit
integration of the underlying differential equation. We then want to
condition the process so that the observable takes a certain value
$F(\eta) = f(\Phi(\eta)) = z\in\RR$. In other words, we want to
restrict ourselves only to noise realizations
$\eta\in L^2([0,T],\RR^m)$ that realize a given value $z\in\RR$ under
the process~(\ref{eq:SDE}).

While we could consider conditioning on more than a scalar variable,
i.e.~$z\in\RR^d$, we will restrict ourselves to the scalar case for
ease of notation. The vectorial case is conceptually the
same. Similarly, we could (and in fact will, for example in
section~\ref{sec:stochastic-levy-area}) consider multiplicative
Gaussian noise, by taking a $\sigma(X_t)$ dependent on the state $X_t$
of the process, which leads to some additional terms further on, but
is, again, conceptually the same.

\section{Pathspace Metropolis-Hastings on conditioned subset}
\label{sec:pathsp-metr-hast}

The goal is now to \emph{sample} the process~(\ref{eq:SDE})
conditioned on the outcome $z$, i.e.~drawing trajectories $\phi(t)$
solution to~(\ref{eq:phi-map}) with $f(\phi)=z$, but with correct
weights, in that more likely ways to realize the constraint should be
preferred over less likely ones. This will yield an \emph{ensemble} of
conditioned trajectories, which is helpful for analyzing physical
mechanisms that are responsible for realizing the given
observable. For example, conditioning a nonlinear wave equation on a
high wave at some final time, we might be interested in investigating
\emph{how} such large waves are produced by drawing many samples of
them and investigating their common features. If the event is rare,
this might otherwise be prohibitively expensive to do. Being able to
sample the conditioned process is even more important when considering
obtaining conditioned probabilities or probability distributions in
complicated physical processes. For example, for a weather model, one
might be interested in the probability distribution of wind speeds at
a given location, conditional on there being anomalously high cloud
cover for a long duration, for the purposes of estimating necessary
energy storage capacities for renewable energy.

\subsection{Bridge Sampling and Pathspace Langevin MCMC}
\label{sec:pathsp-lang-mcmc}

A special case is the situation where we choose $f(\phi) = \phi(T)$,
i.e.~where the observable of interest is the endpoint itself. Since
the state space is $\RR^n$, this is a conditioning on more than a
scalar constraint if $n>1$, but related to our setup. In particular,
this is the case of \emph{diffusion bridge sampling} or
\emph{transition path sampling}.

We first discuss the case where the diffusion vector field is
gradient, $b(x) = -\nabla V(x)$, for a potential $V:\RR^n\to\RR$,
which is of considerable interest in for example molecular dynamics,
where one is interested in a \emph{reaction} as sample path connecting
two local minima of a free energy landscape. Then, simplifications
apply, such that transition path sampling can be implemented very
efficiently. The reason for that is first that knowledge of the
potential landscape implies knowledge of the invariant measure of the
process, and second that gradient dynamics correspond to a
\emph{reversible} Markov process, such that the time-reversed dynamics
can feasibly be integrated as well. As such, transition paths fulfill
time-reversibility constraints, which allow for more effective
sampling~\cite{ bolhuis-chandler-dellago-etal:2002,
  e-ren-vanden-eijnden:2005, bolhuis-swenson:2021}.

In the non-reversible setup, i.e.~for arbitrary vector fields
$b:\RR^n\to\RR^n$, the situation is less straightforward. Still, in
general, we know that the Onsager-Machlup functional
\begin{equation*}
  S_{\text{OM}}(\phi) = \tfrac12\int_0^T \left(\|\dot\phi-b(\phi)\|_\sigma^2 + \nabla\cdot b(\phi)\right)\,dt
\end{equation*}
(for a specific $\sigma$-dependent norm $\|v\|_\sigma = (v
\cdot (\sigma\sigma^\top)^{-1} v)^{1/2}$) quantifies a
formal ``density'' in pathspace of sample paths of the
process~(\ref{eq:SDE}), or in other words, the pathspace measure
formally fulfills $d\mathbb P(\phi) \sim e^{-S_{\text{OM}}(\phi)}
d\phi$.

Therefore, we can build a Langevin MCMC algorithm by constructing a
Markov chain with invariant measure proportional to
$e^{-S_{\text{OM}}}$, which is given for example by the pathspace
Langevin equation in virtual (``algorithmic'') time $\tau$ given by
\begin{equation}
  \label{eq:SPDE}
  \partial_\tau \phi(t,\tau)= -\frac{\delta S_{\text{OM}}(\phi)}{\delta \phi} +
  \xi(t,\tau)\,,
\end{equation}
where $\EE \xi\xi' = \delta(t-t')\delta(\tau-\tau')$ is white-in-time
and virtual time.  When expanding the variational derivative of the
Onsager-Machlup action, the SPDE~(\ref{eq:SPDE}) for $\phi(t,\tau):
[0,T]\times\RR^+\to\RR^n$, then becomes,
\begin{equation}
  \label{eq:SPDE-full}
  \partial_\tau \phi = (\sigma\sigma^\top)^{-1} \partial_{tt} \phi - (\sigma\sigma^\top)^{-1}\nabla b(\phi) - \nabla b(\phi)^\top (\sigma\sigma^\top)^{-1} - b(\phi)\nabla b(\phi) - \nabla (\nabla\cdot b(\phi)) + \xi(t,\tau)
\end{equation}
equipped with boundary conditions
\begin{equation}
  \label{eq:SPDE-BC}
  \phi(0,\tau)=x_0\qquad \text{and}\qquad \phi(T,\tau)  = z\,.
\end{equation}
The SPDE~(\ref{eq:SPDE-full}) has as unique invariant measure (in
pathspace) the bridge measure of~(\ref{eq:SDE}) conditioned on
$\phi(T)=z$. While these arguments are rather formal, they can be made
rigorous for a variety of cases~\cite{stuart-voss-wilberg:2004,
  hairer-stuart-voss:2007, beskos-roberts-stuart-etal:2008}, though
notably still appear to be open for arbitrary vector fields $b$ (since
then the SPDE~(\ref{eq:SPDE-full}) might become singular). This
pathspace Langevin MCMC sampler can even be imbued with
metadynamics~\cite{bussi-laio:2020} to sample in the case of multiple
coexistent transition channels~\cite{grafke-laio:2024}.

Unfortunately, constraints different from mere endpoint constraints
can no longer be implemented in the form of~(\ref{eq:SPDE}) in an easy
way: We would need to draw new paths $\phi$ according to more
complicated restrictions, for example on whole integrals of the
trajectory, which are not simply encoded in a physical-time boundary
condition as in~(\ref{eq:SPDE-BC}). The same complications arise if
$\sigma$ is not invertible, which is generally true if $m<n$,
as~(\ref{eq:SPDE-full}) relies on the existence of
$(\sigma\sigma^\top)^{-1}$. This case, sometimes called the
\emph{degenerate} or \emph{hypoelliptic} case, is often relevant in
practical application as not all degrees of freedom of a physical
system are presumed to be uncertain or random. The fact that in this
case the bridge sampling fails is rather intuitive, since it just
means that the noise-to-path map $\Phi(\eta)$ is not invertible (in
that there are paths $\phi$ that cannot be realized by any choice of
noise $\eta$). In this case, one would rather want to draw random
noise realizations (which are only conditioned on a specific
observable outcome) as opposed to random trajectory realizations
(which must be realizable by the noise). This is indeed the direction
that we will follow here.

\subsection{Conditioned sampling as sampling on a manifold}
\label{sec:cond-sampl-as}

Instead, we might consider sampling not for trajectory realizations
$\phi$, but employing the same idea for noise realizations
$\eta$. Formally, we can consider $\eta ``\!\sim\!" dW/dt$
in~(\ref{eq:phi-map}), i.e.~consider a measure $\mu$ on the space of
trajectories that scales like
\begin{equation}
  \label{eq:formal-noise-prob}
  d\mu(\eta) \propto \exp\left(-\tfrac12\int_0^T |\eta(t)|^2\,dt\right) d\eta = \exp(-\tfrac12\|\eta\|^2) d\eta = p(\eta)\,,
\end{equation}
where the norm $\|\cdot\|$ is $L^2([0,T],\RR^m)$ (this intuitive
notion can be made precise via the classical Wiener space, as we
discuss in appendix~\ref{sec:metr-hast-wien}). While it is not
difficult to draw noise realizations $\eta$, the problem is that
almost surely a random $\eta$ will not fulfill our constraint
$F(\eta)=z$.  For this reason, the main idea we consider here is to
define a manifold in the space of noise realizations
\begin{equation*}
  \mathcal M_z = \{ \eta\in L^2([0,T],\RR^m)\ |\ F(\eta)=z\}\subset L^2([0,T],\RR^m)\,,
\end{equation*}
consisting of noise realizations that fulfill the global constraint on
$\Phi(\eta)$ (i.e.~$\mathcal M_z = F^{-1}(\{z\})$). In other words, we
want to sample a Gaussian random variable on a submanifold. This can
be achieved by drawing noise realizations under the constraint for
example via an MCMC with an update step and subsequent projection onto
$\mathcal M_z$.

\citet{zappa-holmes-cerfon-goodman:2018} specify a scheme to sample a
probability measure known up to normalization constant, but
constrained onto a submanifold of Euclidean space (see
also~\cite{girolami-calderhead:2011, lelievre-rousset-stoltz:2012,
  lelievre-rousset-stoltz:2019, xu-holmes-cerfon:2024} for sampling on
submanifolds, and~\cite{hairer:2000} for solving differential
equations on constraint manifolds).  The main idea is to construct a
Metropolis adjusted Markov chain Monte Carlo on manifolds by a careful
(and statistically reversible) projection procedure. Concretely, given
a point $x\in \mathcal M_z$, we find a new point $y\in \mathcal M_z$
by the following steps:
\begin{enumerate}[(i)]
\item Find the tangent space $T_x \mathcal M_z$.
\item Draw a random element $v\in T_x \mathcal M_z$ from this tangent
  space, for example normally distributed.
\item In general, $x+v\notin \mathcal M_z$, and we need to project
  back onto $\mathcal M_z$. This is done by finding a $w \perp T_x
  \mathcal M_z$ such that $x + v + w = y \in \mathcal M_z$.
\item Accept or reject $y$ based on a Metropolis Hastings rejection
  step to achieve the desired target distribution.
\end{enumerate}
We argue here that the above remains not only valid, but numerically
practical, when applied to whole trajectories, i.e.~to sampling noise
realizations $\eta\in L^2([0,T],\RR^m)$ defining trajectories of
stochastic differential equations, constrained to a manifold that
realizes a scalar constraint $F(\eta)=z$. In the following, we will
discuss the above steps in light of this generalization.

\subsection{Finding the tangent space}
\label{sec:find-tang-space}

In order to find the tangent space $T_\eta \mathcal M_z$ at a given
noise realization $\eta$, we require the computation of the gradient
of the noise-to-observable map $F = f\circ \Phi$. Given that $F$
implicitly necessitates the solution of an ODE and gradients are
potentially high dimensional, care must be taken to compute it in a
scalable way.  We can apply the adjoint formalism (following for
example~\cite{plessix:2006}) to our problem of computing the normal
vector to $\mathcal M_z$ at a point $\eta\in L^2([0,T],\RR^m)$. To do
so, define the forward equation implicitly by constructing a residual
map $\Psi: L^2([0,T],\RR^n)\times L^2([0,T],\RR^m) \to
L^2([0,T],\RR^n)$ with
\begin{equation*}
  \Psi(\phi,\eta) = \partial_t \phi - b(\phi) - \sigma \eta
\end{equation*}
so that $\Psi(\phi,\eta)=0$ whenever the trajectory and noise pair
$(\phi,\eta)$ fulfill our original SDE. Then, we can set
\begin{equation*}
  \frac{\delta F}{\delta \eta} = \left(\frac{\delta \Psi}{\delta \eta}\right)^\top \mu
  \quad\text{where $\mu$ is given by}\quad
  \left(\frac{\delta \psi}{\delta \phi}\right)^\top \mu = \frac{\delta f}{\delta \phi}\,.
\end{equation*}
In particular, it is clear that
\begin{equation*}
  \frac{\delta \Psi}{\delta \phi} = \partial_t - \nabla b(\phi)
  \quad\text{and}\quad
  \frac{\delta \Psi}{\delta \eta} = \sigma\,,
\end{equation*}
so that we have $\frac{\delta F}{\delta \eta} = \sigma^\top \mu$ where
$\mu$ solves the differential equation
\begin{equation}
  \label{eq:adjoint}
  \partial_t \mu = -\nabla b(\phi)^\top \mu - \frac{\delta f}{\delta \phi}\,.
\end{equation}
Equation~(\ref{eq:adjoint}) is the \emph{adjoint equation} to the
forward equation~(\ref{eq:phi-map}). We have now found the normal
vector at $\eta$ to $\mathcal M_z$: It is the vector $n_\eta =
\sigma^\top \mu \in L^2([0,T],\RR^m)$ obtained by solving the adjoint
equation~(\ref{eq:adjoint}). Even though we need to keep in mind that
$\eta$ is not differentiable in time, and so care must be taken to define
the proper derivative in terms of stochastic analysis (compare
appendix~\ref{sec:cond-expect}), ultimately the finite dimensional
truncation remains the same.

This also allows us to project onto the tangent space $T_\eta \mathcal
M_z$ of $\mathcal M_z$ at location $\eta$ via the orthogonal
projection operator
\begin{equation*}
  \Pi_{T_\eta \mathcal M_z} v = v - \frac{\langle n_\eta,
    v\rangle}{\langle n_\eta, n_\eta\rangle} n_\eta\,,
\end{equation*}
where the inner product $\langle\cdot,\cdot\rangle$ is in
$L^2([0,T],\RR^m)$. This projection operator can be used to generate
proposed update vectors $v\in T_\eta\mathcal M_z$, for example by
drawing a normally distributed vector, and applying $\Pi_{T_\eta
  \mathcal M_z}$ to it. We now want to discuss a few special cases of
observables.

\subsubsection{Endpoint constraints}

Often, we want a constraint on the endpoint, i.e.~we want to constrain
the final point $X_T$ for the SDE~(\ref{eq:SDE}) to be equal to a
value $z\in\RR$. This is the case of the stochastic bridge. In this
case, the observable $f(\phi) = g(\phi(T))$, so that formally
$\frac{\delta f}{\delta \phi} = \delta(t-T) (\nabla g(\phi(T)))$. This
then yields a \emph{boundary condition} for the adjoint
equation~(\ref{eq:adjoint}), i.e.
\begin{equation*}
  \partial_t \mu = -\nabla b(\phi)^\top \mu\,,\qquad \mu(T) = -\nabla g(\phi(T))\,.
\end{equation*}

\subsubsection{Integral constraints}

Instead, we might want to have an integral constraint, i.e. $f(\phi) =
\int_0^T g(\phi(t))\,dt$. In that case, $\frac{\delta f}{\delta \phi}
= \nabla g(\phi(t))$, which yields a source term for the adjoint
equation,
\begin{equation*}
  \partial_t \mu = -\nabla b(\phi)^\top \mu -\nabla g(\phi(t))\,,\qquad \mu(T)=0.
\end{equation*}

\subsection{Backprojection onto the submanifold}

\begin{figure}
  \begin{center}
    \includegraphics[width=0.3\textwidth]{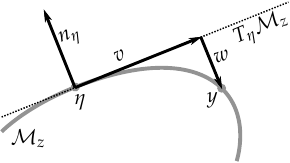}
  \end{center}
  \caption{Sketch of the submanifold sampling technique. Given a point
    $\eta\in\mathcal M_z$, and a random $v\in T_\eta\mathcal M_z$, find
    $w\perp T_\eta\mathcal M_z$ such that $y=\eta+v+w\in\mathcal
    M_z$. Lastly, the proposal $y$ is accepted via a
    Metropolis-Hastings rejection step.}
  \label{fig:manifold}
\end{figure}

Once a $v\in T_\eta \mathcal M_z$ is found, of course in general
$\eta+v\notin \mathcal M_z$. We therefore must additionally
backproject $\eta + v$ onto $\mathcal M_z$, by finding a $w\parallel
n_\eta$ such that $\eta+v+w\in\mathcal M_z$. This can be achieved, for
example, by a Newton solver, to solve $F(\eta+v+\alpha n_\eta)$ for
$\alpha\in\RR$. Crucially, not only does the Newton solver converge to
acceptable accuracy extremely quickly, but it also does need only
first derivatives of $F$, which we already know how to compute, and no
higher derivatives are required. As discussed
in~\cite{zappa-holmes-cerfon-goodman:2018}, this is in contrast to a
possibly more intuitive \emph{orthogonal projection} onto $\mathcal
M_z$. This whole procedure is depicted in figure~\ref{fig:manifold}.

\subsection{Metropolis Hastings adjustment}
\label{sec:backpr-onto-mathc}

Now, given a new candidate $\eta + v + \alpha n_\eta = y\in\mathcal
M_z$, we need to decide whether to accept or reject this proposal
point. This is achieved by taking an acceptance probability of
\begin{equation*}
  \mathrm{accept}(y|\eta) = \min\left(1, \frac{p(y) q(v'|y)}{p(\eta) q(v|\eta)}\right)\,.
\end{equation*}
Here, $p(\eta)$ and $p(y)$ are given by~(\ref{eq:formal-noise-prob})
restricted to $\mathcal M_z$ (compare appendix~\ref{sec:cond-expect}),
while $q(v|\eta)$ is the probability of drawing the update vector
$v\in T_\eta \mathcal M_z$ (which we usually choose independent of
$\eta$), and in reverse, $q(v'|y)$ is the probability to have drawn
$v'\in T_y\mathcal M_z$ for a step in the opposite direction. For
this, note that $v' = \Pi_{T_y \mathcal M_z} (\eta-y)$, which
necessitates computing the normal vector at $y$. Of course, if the
proposal ends up being accepted, this normal vector can be reused for
the next iteration of the algorithm.

It is important to note that on top of the ratio of probabilities of
the points, and the ratio of conditional probabilities of the updates,
an additional factor must be considered in order to ensure
reversibility of the Monte Carlo algorithm: We need to ensure that
given $y$ and $v'\in T_z\mathcal M_z$, that a Newton search starting
from $y+v'$ in direction $n_y$ would end up in the original point $\eta$
again. If this is not the case, we need to reject the proposal.

Lastly, while the stochastic noise for the original process is drawn
from a Gaussian measure on continuous functions (i.e.~the Wiener
measure), we are actually interested in sampling only from the measure
restricted to the submanifold $\mathcal M_z$, so that all samples
fulfill the constraint $F(\eta)=z$. Since the manifold $\mathcal M_z$
is specified as isoset of the function $F$, there are geometric
considerations to be taken into account as to how the parametrization
of $F$ influences the measure restricted to $\mathcal M_z$. As
discussed in appendix~\ref{sec:cond-expect}, this is given by the
coarea formula, which assigns to the acceptance probability an
additional ratio of volume elements due to the parametrization of
$\mathcal M_z$.

\subsection{Metropolis Hastings on Hilbert spaces}

Even without any constraints, sampling paths of a continuous-time
stochastic process such as an SDE implies that we are dealing with
infinite dimensional Hilbert spaces. While the above arguments are
sound for finite dimensional vector spaces, and one might correctly
argue that they therefore also apply to any truncated numerical
discretization, it has been
noted~\cite{cotter-roberts-stuart-etal:2013, kantas-beskos-jasra:2014,
  cui-law-marzouk:2016, beskos-girolami-lan-etal:2017,
  rudolf-sprungk:2018, garbuno-inigo-hoffmann-li-etal:2020} that the
naive extension to the functional setup comes with scaling problems if
not carefully treated. Concretely, if we choose to discretize the time
interval $[0,T]$ with $N_t$ time steps, then the acceptance
probability of a Metropolis-Hastings sampler generally approaches zero
as $N_t\to\infty$. The continuum limit $N_t\to\infty$ in that sense
can be viewed as doubly problematic: All computations discussed above,
such as the computation of the gradient, or the backprojection onto
the manifold, necessitate the solution of differential equations, the
cost of which naturally grows when more timesteps are
considered. Generally, this is expected to scale linearly with the
sample space dimension $N_t$. In addition to this, we need to consider
the number of necessary iterations of the algorithm until it is
well-mixed or has sufficiently explored the configuration space, and
its scaling with the sample space dimension. It has been realized
in~\cite{cotter-roberts-stuart-etal:2013} and rigorously proven
in~\cite{hairer-stuart-vollmer:2014} that in special cases a careful
construction of the proposal transition kernel allows the design of
Metropolis-Hastings algorithms for which the number of necessary steps
in order to reach a given accuracy is \emph{independent} of the sample
space dimension. In general, though, the question of scalability of
algorithms of the above type is a difficult one. In
appendix~\ref{sec:metr-hast-wien}, we will briefly summarize results
for MCMC on Hilbert spaces, and highlight the associated
problems. While these arguments become fairly technical rather
quickly, to our knowledge the dimension-independent scaling of MCMC on
arbitrary manifolds is an unsolved problem. As we demonstrate in the
examples section that follows, though, in many cases of practical
interest the disadvantageous scaling is still acceptable and provides
usable results, even though the high-dimensional limit poses
limitations.

\section{Examples}
\label{sec:examples}

As examples, we will consider applications from various different
areas: First, as demonstration that the algorithm reproduces
well-known results, we will sample well-known properties of Brownian
bridges in section~\ref{sec:range-stat-stand}. Subsequently, in
section~\ref{sec:dynam-cond-phase}, we will sample a dynamical
condensation phase transition as a case of conditioning on a temporal
integral. Drawing sample paths realizing a given large stochastic Levy
area, as described in section~\ref{sec:stochastic-levy-area} is an
example conditioning on a \emph{stochastic} integral. The last two,
namely conditioning the Korteweg-deVries equation for shallow water
waves on large wave outcomes in section~\ref{sec:pde-with-endpoint},
and conditioning a model of turbulence in subcritical pipe flow on
relaminarization in section~\ref{sec:decay-turb-puffs} constitute two
examples of stochastically perturbed \emph{partial} differential
equation conditioned on an endpoint constraint, demonstrating that the
discussed algorithm remains feasible for very large dimensions.

\subsection{Range statistics of the standard Brownian bridge}
\label{sec:range-stat-stand}

\begin{figure}
  \begin{center}
    \includegraphics[width=0.48\textwidth]{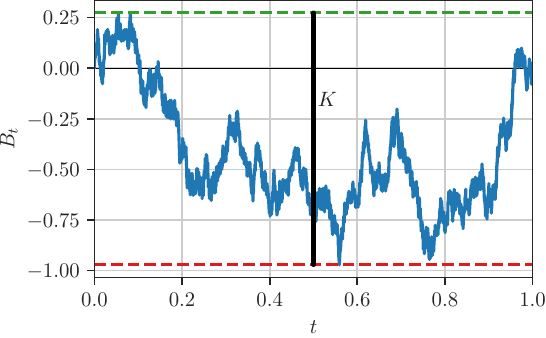}
    \includegraphics[width=0.48\textwidth]{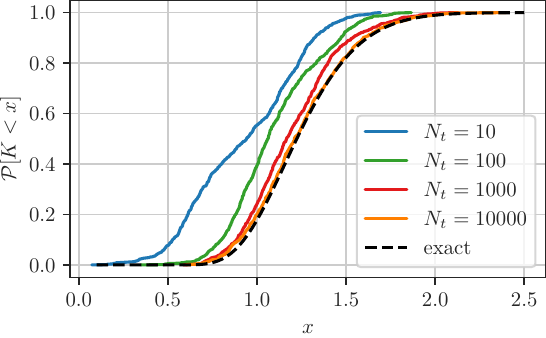}
  \end{center}
  \caption{Range statistics of the standard Brownian bridge. Left: For
    a standard Brownian bridge (blue), the range $K$ is given by the
    difference between its maximal (green) and minimal (red)
    point. Right: The known statistics of $K$ given
    by~(\ref{eq:range-statistics}) (black dashed) are reproduced by
    taking empirical statistics over conditioned paths produced by our
    algorithm (colored lines).}
  \label{fig:brownian-range}
\end{figure}

Consider, for 1-dimensional Brownian motion $W_t$, the standard
Brownian bridge $B_t$ given by
\begin{equation}
  \label{eq:brownian-bridge}
  B_t = W_t - t W_T\,,\quad B_0=0\,,\quad t\in[0,T]\,, T=1\,.
\end{equation}
While it is not difficult to sample~(\ref{eq:brownian-bridge}), and
the above approach is far too general, the Brownian bridge clearly
falls into the class of conditioned stochastic differential equations
as it can be viewed as a standard Brownian motion conditioned on
hitting $W_T=0$ for $T=1$. To interpret it in our framework, we set
\begin{equation*}
  \Phi :L^2([0,1],\RR) \to L^2([0,1],\RR)\quad\text{with}\quad \Phi: \eta\mapsto \int_0^t \eta(\tau)\,d\tau
\end{equation*}
that maps noise to trajectory, and as observable take
\begin{equation*}
  f(\phi) = \phi(T)\,,
\end{equation*}
so that $\mathcal M_{z=0} = \{ \eta \in
L^2([0,1],\RR)\ |\ \Phi(\eta)(T)=0\}$ is the constraint manifold
corresponding to the standard Brownian bridge. Note that in this case,
the adjoint equation~(\ref{eq:adjoint}) is trivially solved by the
temporally constant $\mu(t) = z$ and hence the constraint manifold
$\mathcal M_z$ is a hyperplane and not curved. This simplifies the
algorithm significantly, since (i) the tangent space is identical
everywhere, (ii) for any $v\in T_x\mathcal M_z$ we have that
$x+v\in\mathcal M_z$ as well, so the projection step is
redundant.

The example nevertheless serves as a helpful test case in order to
ensure that the algorithm produces correctly conditioned sample paths,
both in respect to the scaling limit of taking the number of temporal
discretization points to infinity, as well as regarding the
statistical properties of the sample paths themselves. As a benchmark
test, we want to obtain statistics of the range of the Brownian
bridge, defined as the random variable
\begin{equation*}
  K = \max_{t\in[0,1]} B_t - \min_{t\in[0,1]} B_t\,,
\end{equation*}
i.e.~the difference between the highest and lowest point of the
standard Brownian bridge (see figure~\ref{fig:brownian-range}
(left)). It is known that the cumulative distribution function of $K$
is given by~\cite{perman-wellner:2014}
\begin{equation}
  \label{eq:range-statistics}
  F_K(x) = \mathcal P[K < x] =  \sum_{k=-\infty}^\infty (1-4k^2x^2) \exp(-2k^2x^2)\,.
\end{equation}
Indeed, as visible in figure~\ref{fig:brownian-range} (right), the
correct statistics~(\ref{eq:range-statistics}) (black dashed) are
reproduced by our algorithm for large $N_t$. Note in particular how we
can freely choose very large $N_t$ without suffering mixing penalties
for the high dimension of our sample space if we are using scalable
algorithms. Here, we use preconditioned
Crank-Nicolson~\cite{cotter-roberts-stuart-etal:2013} for suggesting a
proposal, and consequently the acceptance probability of the
Metropolis-Hastings step does not degenerate in the high-dimensional
limit.

\begin{figure}
  \begin{center}
    \includegraphics[width=0.48\textwidth]{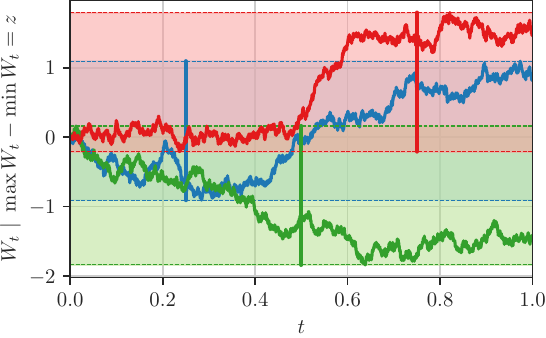}
    \includegraphics[width=0.48\textwidth]{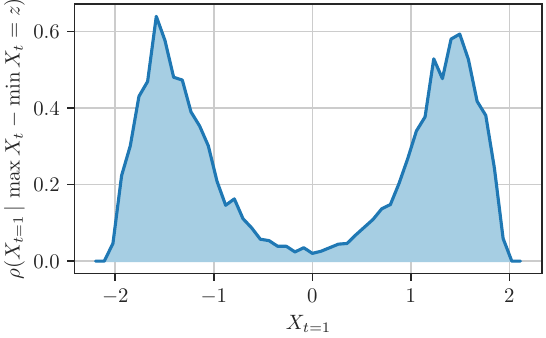}
  \end{center}
  \caption{Brownian motion conditioned on its range $K$. Left: Three
    sample trajectories of Brownian motions $W_t$ such that $\max W_t
    - \min W_t = z=2$. The maximal and minimal value of the sample
    trajectories are indicated by dashed horizontal lines, and the
    range by a solid vertical line. Right: Endpoints $X_{t=1}$ of the
    Brownian motion conditioned on a range of $K=2$ appear to avoid
    small values and are bimodally concentrated around $\pm 1.5$. }
  \label{fig:range-endpoint}
\end{figure}

On the other hand, we are free to look at the exact same problem in
reverse: Taking a standard Brownian motion $X_t = W_t$, we can condition on
its range
\begin{equation*}
  K = \max_{t\in[0,1]} X_t-\min_{t\in[0,1]} X_t
\end{equation*}
and instead look at the statistics of the endpoint $X_1$, for example
the probability distribution function
\begin{equation}
  \label{eq:range-endpoint-pdf}
  \rho(X_1\ |\ K=z)\,,
\end{equation}
for which, to the best of our knowledge, no analytical result is
known. Note that in this case, $\mathcal M_z$ is no longer a
hyper-plane, the tangent space non-trivial, and the algorithm no
longer simplifies. Results are shown in
figure~\ref{fig:range-endpoint}, where on the left we demonstrate for
three sample trajectories that indeed the range $K$ is always fixed
(in this case, we choose $z=2$), while the right shows the empirical
distribution~(\ref{eq:range-endpoint-pdf}) obtained as a histogram
over conditioned paths generated by the algorithm. Notably, this
distribution as expected is symmetric around $0$, and has peaks at
approximately $\pm1.5$, while avoiding the origin. In other words, a
Brownian motion conditioned on a range of $z=2$ will most likely end
away from the origin. We choose $N_t=10000$.

\subsection{Dynamical Condensation Phase Transition in Ornstein-Uhlenbeck process}
\label{sec:dynam-cond-phase}

As a slightly more intricate example, consider the Ornstein-Uhlenbeck
process, for $X_t\in\RR$, given by
\begin{equation}
  \label{eq:ou}
  dX_t = -X_t\,dt + \sqrt{\eps}\,dW_t\,,\qquad X_0=0\,.
\end{equation}
The invariant distribution of this process is Gaussian, with mean $0$
and variance $\eps^2/2$. We now want to condition on the observable
\begin{equation}
  \label{eq:ou-obs}
  f(\phi) = \frac1T\int_0^T \textrm{sign}(\!\phi(t)\!)\,|\phi(t)|^\alpha\,dt\,,
\end{equation}
i.e.~integrating the process to some power $\alpha$, keeping track of
the sign. In particular, for $\alpha=1$, this is simply the temporal
mean of the process, while higher powers $\alpha>1$ can be considered
as well. It is known~\cite{nickelsen-touchette:2018} that this
observable undergoes a condensation phase transition in time with the
order parameter $\alpha$ in the following sense: If $\alpha<2$ then
large values of~(\ref{eq:ou-obs}) are realized by biasing the linear
process~(\ref{eq:ou}) to a new mean, and statistics will be Gaussian
around the new mean. For $\alpha>2$, instead, large values
of~(\ref{eq:ou-obs}) are realized by temporally localized excursions,
while the majority of time the process remains Gaussian with mean
zero. This is reminiscent of a condensation phase transition in
lattice gases, where spatially homogeneous densities transition into
localized density spikes, but here with the time variable replacing
the spatial extend.

\begin{figure}
  \begin{center}
    \includegraphics[width=0.48\textwidth]{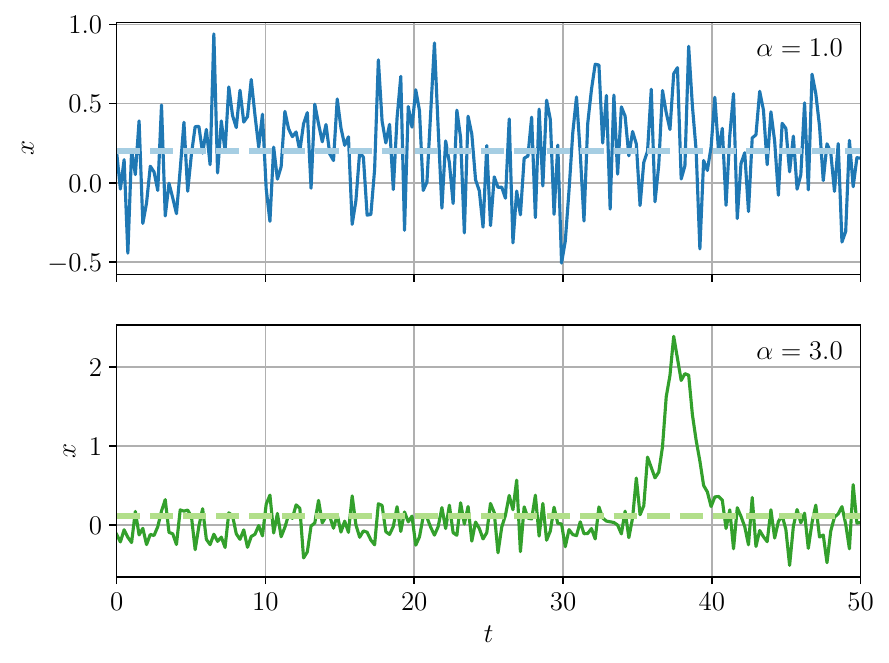}
    \includegraphics[width=0.48\textwidth]{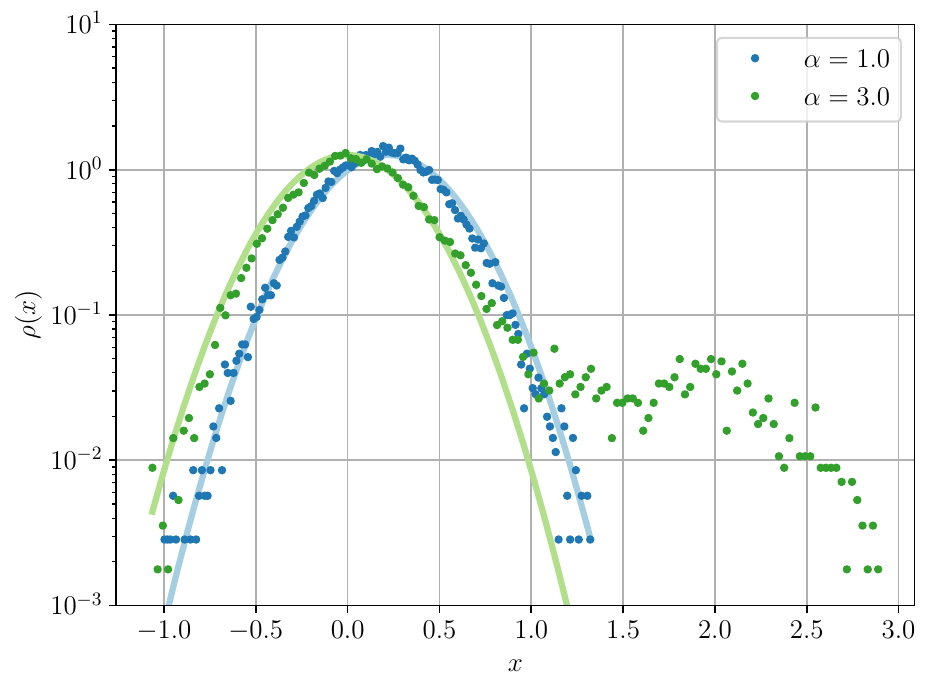}
  \end{center}
  \caption{Dynamical Condensation Phase Transition. Left: Realizations
    of the linear process~(\ref{eq:ou}) conditioned on large
    areas~(\ref{eq:ou-obs}) for different values of $\alpha$. For
    $\alpha=1$ (top), the large average is realized by shifting the
    mean. For $\alpha=3$ (bottom), the large average is realized by a localized
    excursion (dynamical condensate) instead. Right: The statistics
    for $\alpha=1$ (blue), conditioned on large~(\ref{eq:ou-obs}),
    are Gaussian with shifted mean, while for $\alpha=3$ (green) they have a
    mean-zero Gaussian core, but with large right tail that contains
    the dynamical condensate.}
  \label{fig:OU}
\end{figure}

We can probe this phase transition by sampling~(\ref{eq:ou}), but
conditioned on large values of~(\ref{eq:ou-obs}). In
figure~\ref{fig:OU} (left), we see a realization each, for
$\alpha=1.0$ (top) and $\alpha=3.0$ (bottom), both conditioned on
$f(\phi)=z=0.2$, which for the chosen values of $\eps=0.1$, $T=50$, is
a rare outlier event. We see that for $\alpha=1.0$ the event is
realized by the whole process shifting to now fluctuate around the new
mean $0.2$, while for $\alpha=3.0$ the event is realized by a single,
localized excursion (around $t=37$). The same is visible in the
statistics of the process, visualized in figure~\ref{fig:OU} (right):
While the $\alpha=1.0$ (blue) probability distribution of the
conditioned process (shown as markers) remains Gaussian around a new
mean (shown as solid line), for $\alpha=3.0$ (green) the conditioned
probability distribution (markers) has a mean-zero Gaussian core
(solid green line), but an additional heavy tail. We pick $N_t=200$
time steps here.

\subsection{Stochastic Levy area and conditional distributions}
\label{sec:stochastic-levy-area}

\begin{figure}[b]
  \begin{center}
    \includegraphics[height=0.2\textheight]{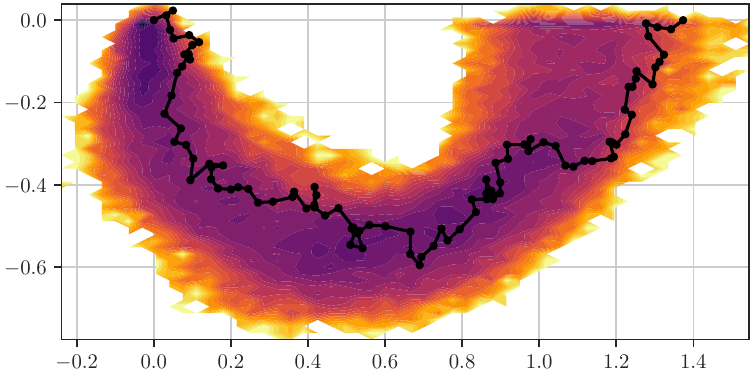}
    \includegraphics[height=0.2\textheight]{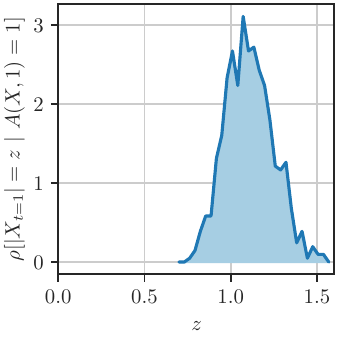}
  \end{center}
  \caption{Left: path density of two-dimensional Brownian motions
    conditioned on a Levy area of 1. Here, for visualization purposes,
    all paths are rotated such that their endpoint is at
    $X^{(2)}_1=0$. Right: Conditional probability distribution for the
    distance of the endpoint from the origin, conditioned on the Levy
    area being equal to one.}
  \label{fig:levy}
\end{figure}

The stochastic Levy area~\cite{levy:1940, levy:1951,
  yor:1980,ikeda-kusuoka-manabe:1995,ferreiro-castilla-utzet:2011,
  buisson-mnyulwa-touchette:2023} is defined as the area enclosed by a
two-dimensional Brownian motion. More precisely, considering the SDE
for $X_t\in\RR^2$ given by
\begin{equation}
  \label{eq:levy-sde}
  dX_t = dW_t\,,\quad X_0 = 0\,,\quad t\in[0,1]
\end{equation}
with $X_t = (X_t^{(1)}, X_t^{(2)})$, the stochastic Levy area is given
by
\begin{equation*}
  A(t,X) = \tfrac12 \int_0^T (X^{(1)}_s\, dX_s^{(2)} - X^{(2)}\, dX_s^{(1)})\,,
\end{equation*}
which is a stochastic integral in the Ito-sense. We now wish to take
as observable the Levy area and condition on a Levy area value of $1$,
arbitrarily. The algorithm then allows us to draw trajectories from
the SDE~(\ref{eq:levy-sde}), but always conditioned on the fact that
its enclosed area is equal to 1. Figure~\ref{fig:levy} (left) depicts,
in $\RR^2$, a density of such paths drawn from the algorithm. Since
obviously the process is rotation symmetric, here for visualization
purposes we choose to rotate all paths such that their endpoint falls
onto the $x$-axis. This allows to visually identify the mechanism by
which the path realizes the given area, since all paths lie in a clear
tube forming roughly a semicircle. In figure~\ref{fig:levy} (right),
we see the conditional probability density of $|X_{t=1}|$ over such
paths, namely
\begin{equation*}
 \rho(|X_{t=1}|>z\ \big|\ A(1,X)=1)\,,
\end{equation*}
i.e.~the the probability density of the distance to the origin of the
endpoint at $t=1$, given that the Levy area is equal to 1.

\subsection{PDE with endpoint constraint: Stochastic Korteweg-deVries equation}
\label{sec:pde-with-endpoint}

Here, we consider a more complicated example by taking as
stochastic PDE the Korteweg-deVries equation for shallow water
waves. For an elevation field $\phi(x,t) : [0,L]\times[0,T]\to\RR$, it
is given by
\begin{equation}
  \label{eq:kdv}
  \partial_t \phi = \kappa\partial_{xxx} \phi + \alpha \partial_{xx} \phi - \phi\partial_x\phi + \eta
\end{equation}
with dispersion coefficient $\kappa>0$, diffusion coefficient
$\alpha>0$, and stochastic forcing $\eta$, which we take to be
long-range correlated in space, white-in-time,
\begin{equation*}
  \EE \eta(x,t)\eta(x',t') = \delta(t-t') \chi(x-x')
\end{equation*}
where in our case $\chi(z)$ only contains the largest-scale Fourier
mode contained in the domain (i.e. corresponding to $k=2\pi/L$).

We are then interested in sampling large waves from a surface at rest
at $t=0$, such that we observe a high wave amplitude at $x=0$ at
$t=T=1$, corresponding to an observable
\begin{equation*}
  f(\phi) = \phi(x=0,t=1)\,.
\end{equation*}
Since only the largest non-constant mode is stochastically forced, the
underlying dimensionality of the sampling problem is not actually
higher than that of a low-dimensional SDE, and in fact the noise is
$\eta \in L^2([0,T],\RR^2)$: Only amplitude and phase of the noise are
random (at each instance in time), while its wave number is
given. Equivalently, we can view the noise as a complex-valued
white-in-time noise. Nevertheless, we need to solve the whole
PDE~(\ref{eq:kdv}) on $[0,T]$ to obtain the surface height at final
time at the origin, such that the computation of gradients, both for
obtaining the tangent space and for back-projecting onto $\mathcal
M_z$, involves deriving through a PDE.

\begin{figure}
  \begin{center}
    \includegraphics[width=0.75\textwidth]{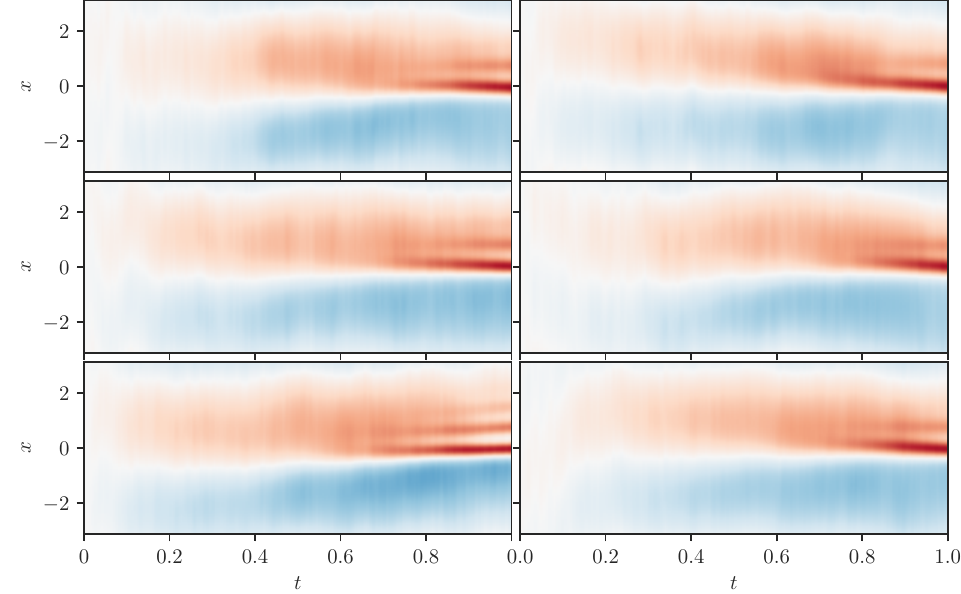}
  \end{center}
  \caption{Six samples of the stochastic Korteweg-deVries
    equation~(\ref{eq:kdv}), conditioned on observing a high wave
    amplitude at $T=1$ at the origin $x=0$. For long times, only
    random fluctuations in the forced largest scale mode are observed,
    until the nonlinearity leads to a steepening of the wave and the
    dispersion then breaks it into multiple wave packets, the largest
    of which becomes the high wave we condition on. The six samples
    show that even though all large waves evolve in a similar way,
    there is considerable variety in their detail such as the number
    of crests or their movement speed.}
  \label{fig:kdv-samples}
\end{figure}

\begin{figure}
  \begin{center}
    \includegraphics[width=0.7\textwidth]{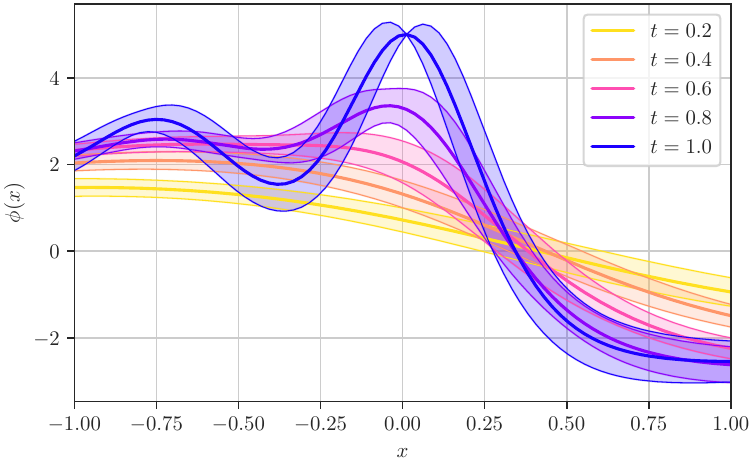}
  \end{center}
  \caption{Evolution of the Korteweg-deVries equation into a large
    wave at $t=1$ at $x=0$. Here, we choose $z=5.0$, and plot an
    ensemble of trajectories that realize this constraint. We
    additionally plot the standard deviation of all observed
    realizations.}
  \label{fig:kdv}
\end{figure}

Figure~\ref{fig:kdv} depicts the result of the algorithm, drawing six
samples of the stochastic KdV equation~(\ref{eq:kdv}), conditioned on
realizing wave height $z=5.0$ at $t=T=1$ at $x=0$, while all other
degrees of freedom are free. The trajectories are shown in space and
time, starting with a flat surface at $t=0$, and evolving such that a
high amplitude wave appears at the origin at final time. There is
considerable variety between the samples, concerning e.g.~the number of
wave crests or the temporal evolution of the zero crossing point.

In figure~\ref{fig:kdv}, we show the mean realization over a sampling
run, where all samples adhere to the constraint. We additionally show
the region of $\pm$ one standard deviation around this mean
realization, signifying the possible deviations of observed large
waves. The standard deviation must be zero at $x=0$, since all
trajectories fulfill the constraint. Generally, the deviations allow
for the wave samples to always look roughly identical, but for example
deviate a bit to the left or right.

\subsection{Decay of turbulent puffs in subcritical pipe flow}
\label{sec:decay-turb-puffs}

The transition to turbulence in subcritical pipe flow is a classical
problem: As first described by~\citet{reynolds:1883}, fluid in a pipe
has a transitional regime, where the laminar flow coexists with
long-lived turbulent structures, so-called ``puffs''. With increasing
Reynolds number $r$, puffs become more and more long-lived, but the
laminar state remains a linearly stable solution throughout. In this
regime, an initially laminar pipe, if left completely undisturbed,
remains laminar indefinitely, while a turbulent puff, once excited, is
extremely long-lived as well~\cite{avila-moxey-lozar-etal:2011}.

The ultimate question whether turbulent flow prevails in the pipe is
therefore connected to the lifetime of turbulent puffs, and the
question of their route to decay back into laminar flow. Since the
puffs at intermediate Reynolds numbers are long lived, their
decay cannot easily be observed or simulated in this regime, and
bespoke rare events algorithms need to be
employed~\cite{rolland:2018}.

Here, instead, we want to condition the fluid in a pipe to its laminar
state in the future, while starting it with a turbulent puff
initially. Sampling this process should then generate samples of
decaying turbulent puffs, which would otherwise be very rare to
observe. To this end, we utilize a stochastic model of turbulent
pipe flow introduced by~\citet{barkley:2016}, given by the PDE
\begin{equation}
  \label{eq:barkley-model}
  \begin{cases}
    \partial_t q +(u-\zeta)\partial_x q= f_r(q,u)+D\partial_x^2q+\sigma q \eta \\
    \partial_t u +u\partial_x u= \epsilon\left[ (U_0-u)+\kappa(\bar{U}-u)q\right]\\
  \end{cases}\,,\quad x\in[0,L]\,, t>0\,,
\end{equation}
with $f_r(q,u)=q(r+u-U_0-(r+\delta)(q-1)^2)$. Here, $x$ is the
direction along the pipe, $u(x,t)$ denotes the mean shear velocity,
and $q(x,t)$ quantifies the turbulent velocity fluctuations away from
the laminar profile. In other words, $q=0$ corresponds to laminar
flow, while a turbulent puff is a localized region where
$q>0$. Conceptually, this model is a two-species
reaction-advection-diffusion equation, but with parameters carefully
fitted to the fluid dynamical situation at hand. The parameter $r$
corresponds to the Reynolds number, and $\eta$ is a spatiotemporal
white noise with amplitude $\sigma$,
\begin{equation*}
  \EE \eta(x,t)\eta(x',t') = \delta(t-t')\delta(x-x')\,.
\end{equation*}
Note that in~(\ref{eq:barkley-model}) the stochastic fluctuations are
multiplicative Gaussian, since they are proportional to $q$, which
makes physical sense: $q$ signifies chaotic excursions of the
Navier-Stokes equations, which are absent for the laminar flow $q=0$
and become stronger with increasing turbulence. 

Equation~(\ref{eq:barkley-model}) is a simplified excitable dynamics
model of turbulent flow, but agrees with phenomena observed in
simulations of the Navier-Stokes equation and experiments to a
remarkable
degree~\cite{barkley-song-mukund-etal:2015,barkley:2016}. From our
perspective, though, it is a formidably complicated example: Not only
is it a stochastic PDE with white-in-spacetime forcing, but
additionally forced degenerately ($u$ has no stochasticity) and
multiplicative, with huge portions of the domain where the noise
disappears, since $q=0$ away from the puff. The higher the Reynolds
number $r$, the longer will a turbulent puff survive, so that
turbulent puff decay is effectively impossible to observe for high
enough $r$. In particular, expected puff lifetimes $\tau$ are
hypothesized and experimentally measured to grow doubly exponentially
with $r$, i.e.~$\tau\sim
\exp(\exp(r))$~\cite{gome-tuckerman-barkley:2022}. Conditioning the
puff onto a decay is therefore a worthwhile endeavor, and getting
statistics of puff decay events very difficult to achieve otherwise.

\begin{figure}
  \begin{center}
    \includegraphics[width=0.75\textwidth]{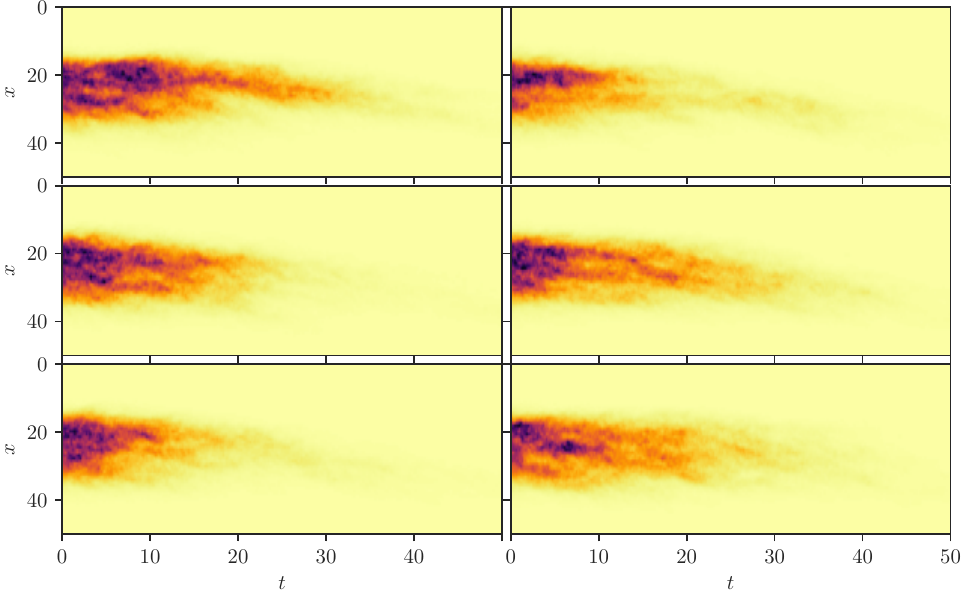}
  \end{center}
  \caption{Six samples of turbulent puff decay in subcritical pipe
    flow for the model~(\ref{eq:barkley-model}). At initial time,
    $t=0$, a turbulent puff is located in the region $x\in[15,30]$. At
    the chosen Reynolds number $r=0.7$, this puff would have an
    extremely long lifetime of more than $10^4$ time units, and
    observing puff decay would be very difficult. We condition it to
    instead laminarize by $T=50$. Depicted is the turbulent velocity
    component $q(x,t)$ along the decay.}
  \label{fig:puffs}
\end{figure}

In our simulations, we use the parameters of~\cite{barkley:2016} at a
Reynolds number $r=0.7$ and noise strength $\sigma=0.25$, with a
periodic domain of size $L=50$. For these parameters, puffs have an
expected lifetime of more than $10^4$ time units: While the decay
event itself happens almost instantaneously on the order of tens of
time units, the fluctuations to trigger it are extremely rare, leading
to long waiting times in the puff configuration before a decay
happens. Concretely, for the application of our algorithm, we define
the \emph{turbulent mass} $Q(t)$ via
\begin{equation*}
  Q(t) = \int_0^L q(x,t)\,dx\,,
\end{equation*}
and condition on $Q(T)=0.1$, i.e.~almost vanishing turbulent mass at
the final time. For comparison, the turbulent mass of a puff is
$Q_{\text{puff}}\approx 25$.

In figure~\ref{fig:puffs}, we show six samples of turbulent puff decay
in space and time: At $t=0$, all samples start with a fully developed
puff, which then gradually decays as time progresses, to an almost
fully relaminarized flow at $T=50$. In an unconditioned run at this
Reynolds number, the expected lifetime of a puff is approximately
$10^4$ time units and puff decay is very rarely
observed. Nevertheless, the decay event itself, once it happens,
typically usually evolves from a viable puff into the laminar flow in
only tens of time units. Physical features of the decay events
depicted in figure~\ref{fig:puffs} qualitatively agree with those of
direct stochastic simulations of the system in the
literature~\cite{barkley:2016, frishman-grafke:2022-a}.

\section{Discussion}
\label{sec:discussion}

We have introduced an algorithm to sample stochastic (partial)
differential equations conditioned on a variety of observables,
including endpoint constraints, constraints on nonlinear temporal
integrals or even stochastic integrals of the process trajectory. The
sampler can be understood as a pathspace Markov chain Monte Carlo
algorithm constrained to a manifold. It allows to directly draw from
the conditioned path measure, thus also enabling access to conditional
distributions of the process, conditioned for example on a rare
event. We have demonstrated the applicability of the algorithm to a
variety of systems, starting from stochastic processes of academic
interest, such as the Brownian bridge or the Levy area, but also
considering physically relevant examples, such as dynamical phase
transitions, ocean surface waves, or supercritical turbulence in
wall-bounded flows. In all cases, we are able to draw from the
conditioned path measure, for example to investigate physical causes
or typical trajectories for the (often rare) event under
consideration.

We remark that a larger number of constraints than a mere scalar
variable is possible without much modification. The number of
constraints is equal to the co-dimension of the submanifold $\mathcal
M_z$, i.e.~if $f:\RR^n \to \RR^d$, then $z\in\RR^d$ and $\mathcal M_z$
is co-dimension $d$. In fact, \citet{zappa-holmes-cerfon-goodman:2018}
treat this case as well: The difference is that the projection step
now needs to search not for a scalar $\alpha\in\RR$ in the Newton
solver step above, but for a coefficient vector $\alpha\in\RR^d$ to
find back to the manifold $\mathcal M_z$. Similarly, projecting onto
the tangent space $T_x\mathcal M_z$ involves removing $d$ components
from an arbitrary proposal vector. Since for systems considered here
generally $d$ is much smaller than the dimensionality of the
surrounding space (which is technically infinite dimensional, or
discretizes to very high dimensions), it is disadvantageous to find an
explicit basis for the tangent space, but projection should remain
feasible.

Additionally, for acceptable convergence and mixing properties, we
rely on the noise-to-observable map $F$ to be sufficiently
nice. Clearly, if the encoded differential equation is very sensitive
to the stochastic input, local curvature of the manifold $\mathcal
M_z$ will be very high, necessitating small update steps to obtain
non-vanishing acceptance rates in light of the projection step. This
might render chaotic systems over long timescales infeasible to be
treated. Similarly, the manifold, even if locally relatively weakly
curved, might intersect with more likely regions multiple times (or
even infinitely many times), leading to a multimodal distribution on
the manifold that is hard to sample. The intuition is that a certain
observable value might be realizable via multiple discretely different
physical mechanisms, and a sampler relying on local updates will
remain stuck in local optimum for a long time. This problem is common
to many other sampling scenarios and can be alleviated in multiple
ways such as replica exchange~\cite{swendsen-wang:1986}, parallel
tempering~\cite{marinari-parisi:1992}, or
metadynamics~\cite{bussi-laio:2020}, all of which could be applied to
the situation at hand (and in part been~\cite{grafke-laio:2024}).

There are obvious and fundamental questions concerning the rigorous
stochastic analysis of the algorithm when considering sampling from
infinite dimensional Hilbert spaces. While we address the basics in
appendix~\ref{sec:metr-hast-wien}, ultimately the problem of
dimension-independent scaling of the sampling algorithm is currently
unsolved, and we need to resort to statements about the scheme
remaining feasible empirically even in the presence of degenerate
scaling. These questions of stochastic analysis become even more
unbearable when considering SPDEs with white-in-space stochasticity,
as we did in section~\ref{sec:decay-turb-puffs}. Here, there is not
even an accepted interpretation of the rigorous meaning of the stated
equation~(\ref{eq:barkley-model}), where distribution-valued rough
fields get freely multiplied in nonlinear terms, without even speaking
about conditioning it on nonlinear observables. Nevertheless, the
model is widely employed and of practical use in the physical
literature (and many others like it in other fields), so that sampling
from its finite-dimensional truncation is of value even if the
continuum limit will remain nebulous for the foreseeable future.

\subsection*{Acknowledgments}

The author wants to thank Timo Schorlepp, Georg Stadler, Jonathan
Weare, Alexandros Beskos, Matthew Graham, and Tim Sullivan for helpful
discussion.

\appendix

\section{Metropolis-Hastings on Wiener space and conditional expectations}
\label{sec:metr-hast-wien}

\subsection{Classical Wiener space}
\label{sec:class-wien-space}

Let $(\mathcal H, \langle\cdot,\cdot\rangle)$ be a separable Hilbert
space, and induced norm as $\|\cdot\|$. For a positive-definite
self-adjoint operator $\mathcal C$, we also define
$\langle\cdot,\cdot\rangle_{\mathcal C} = \langle \mathcal
C^{-1/2}\cdot,\mathcal C^{-1/2}\cdot\rangle$, with induced norm
$\|\cdot\|_{\mathcal C}$.

For the construction of the classical Wiener space, we pick $\mathcal
C = \partial_{tt}^{-1}$ and take $\mathcal H$ to be the space of real
valued functions $h(t)$ on $[0,T]$ with $h(0)=0$ such that
$\|h\|_{\mathcal C}<\infty$. Additionally, we consider the Banach
space $\mathcal B$ of real valued continuous functions equipped with
supremum norm. The classical Wiener measure $\gamma$, i.e.~the law
that defines the process of Brownian motion $W_t$, is a Gaussian
measure on $\mathcal B$ with covariance operator $\mathcal C$, and
$\mathcal H$ is its Cameron-Martin space~\cite{malliavin:2015}.

Formally, this can be interpreted that the density of $\gamma$ is
given via
\begin{equation*}
  d\gamma(h) \sim \exp\left(-\tfrac12 \int_0^T (\partial_t h(t))^2\,dt\right) dh = \exp(-\tfrac12 \|h\|^2_{\mathcal C}) dh\,,
\end{equation*}
with respect to the non-existent Lebesgue-measure $dh$ on $\mathcal
B$. While this is only a formal statement, similar statements with
respect to a common dominating measures make sense, and are relevant
for the formalization of Metropolis-Hastings on Hilbert
spaces~\cite{tierney:1998}.

\subsection{Metropolis-Hastings}

Consider now $\mu$ a target measure on $\mathcal H$ that we want to
sample from, its definition relative to the Gaussian measure $\gamma$
given via $\frac{d\mu}{d\gamma}(x) \propto \exp(-V(x))$ for
convenience for a potential $V$
(compare~\cite{cotter-roberts-stuart-etal:2013,
  hairer-stuart-vollmer:2014}. The idea of the Metropolis-Hastings
algorithm is to construct a Markov process on $\mathcal H$ with
transition kernel $P(x,dy)$ that is reversible with respect to $\mu$,
such that drawing from $\mu$ is equivalent to sampling the Markov
process. Following~\cite{tierney:1998}, to construct $P(x,dy)$, we can
first consider a Markov chain of our choice with transition kernel
$Q(x,dy)$ which we use to generate a proposal. We will accept this
proposal with probability $\alpha(x,y)$. This amounts to a transition
kernel
\begin{equation}
  \label{eq:MH-kernel}
  P(x,dy) = Q(x,dy) \alpha (x,y) + \delta_x(dy)\int (1-\alpha(x,u))Q(x,dt)\,.
\end{equation}
Here, $\alpha$ needs to be chosen such that $P(x,dy)$ is reversible
with respect to $\mu$, i.e.~such that the detailed balance condition
\begin{equation}
  \label{eq:detailed-balance}
  P(x,dy) \mu(dx) = P(y,dx)\mu(dy)
\end{equation}
is fulfilled. For that, define $\nu(dx,dy)=\mu(dx)Q(x,dt)$ and
$\nu^\perp(dx,dy)=\nu(dy,dx)$. Then, there exists a measurable
$R\subset \mathcal H\times \mathcal H$ such that $\nu$ and $\nu^\perp$
are equivalent on $R$ (and mutually singular on its complement). We
can use this to define $r(x,y) = d\nu_{|R}/d\nu_{|R}^\perp(x,y)$ for $\nu$
and $\nu^\perp$ restricted to $R$. It can then be
shown~\cite{tierney:1998} that the detailed balance
condition~(\ref{eq:detailed-balance}) is fulfilled for the
Metropolis-Hastings kernel~(\ref{eq:MH-kernel}) if we choose
\begin{equation}
  \label{eq:acceptance}
  \alpha(x,y) = \begin{cases} \min\{1,r(y,x)\} & \text{if}\ (x,y)\in R\\
    0 & \text{otherwise}\,.
  \end{cases}
\end{equation}
Intuitively, we can interpret $(x,y)\in R$ to mean that $Q(x,dy)$
allows us to reach $y$ from $x$ and vice-versa, and $\alpha(x,y)$ in
(\ref{eq:acceptance}) is chosen precisely such that inserting it
into~(\ref{eq:MH-kernel}) yields~(\ref{eq:detailed-balance}) on $R$:
the diagonal $\delta$-terms cancel for all $\alpha$,
and~(\ref{eq:acceptance}) fixes the off-diagonal.

\subsection{Scaling properties}

While in principle the above formalism is correct for any separable
Hilbert space, including Wiener space, and generalizes to infinite
dimensions, there is a practical
problem~\cite{hairer-stuart-vollmer:2014}. Taking for example a
proposal distribution $y\sim\mathcal N(x,\delta \,\mathcal C)$ (which constitutes
a standard random walk Metropolis), we have that
\begin{equation*}
  \frac{d\nu^\perp}{d\nu}(x,y) = \exp\left(-\tfrac12\left(\|y\|_{\mathcal C}^2 - \|x\|_{\mathcal C}^2\right)\right)\exp\left(V(y)-V(x)\right)
\end{equation*}
independent of $\delta$. Unfortunately, $\|y\|_{\mathcal C}$ is almost
surely infinite for $y\in\mathcal B$, and thus the acceptance
probability is zero. While in practice one can argue that all
numerical approximations of pathspace quantities will end up finite
dimensional, with a discrete number of timesteps $N_t$, this
nevertheless means a broken \emph{scaling} of the algorithm in that
$\alpha(x,y)\to0$ as $N_t\to\infty$. As remarked
in~\cite{cotter-roberts-stuart-etal:2013} and rigorously shown
in~\cite{hairer-stuart-vollmer:2014}, one can circumvent this scaling
problem by a suitable choice of proposals (i.e.~of $Q(x,dt)$).

To that end, and following~\cite{cotter-roberts-stuart-etal:2013}, we
realize that the for a potential $U:\mathcal H\to \RR$, the Markov
process given by the preconditioned Ornstein-Uhlenbeck process in
algorithmic time $\tau$, given by
\begin{equation}
  \partial_\tau h_\tau = \mathcal K \nabla U(h_\tau) + \sqrt{2\mathcal K} \xi_\tau\,,
\end{equation}
with $\mathcal K$ positive definite preconditioning operator, has an
invariant measure proportional to $\exp U(h)$ for any choice of
$\mathcal K$, and where $\xi_\tau$ is $\mathcal H$-valued white noise
in algorithmic time $\tau$. For the case of Gaussian measures with
covariance operator $\mathcal C$, we have $U(h) = -\tfrac12
\|h\|^2_{\mathcal C}$, and hence
\begin{equation}
  \partial_\tau h_\tau = -\mathcal K \mathcal C^{-1} h_\tau + \sqrt{2\mathcal K} \xi_\tau\,,
\end{equation}
which becomes, for the choice $\mathcal K = \mathcal C$,
\begin{equation}
  \label{eq:SPDE-langevin}
  \partial_\tau h_\tau = -h_\tau + \sqrt{2C} \xi_\tau\,,
\end{equation}
with $\xi_\tau\sim \gamma$. We can construct a
proposal scheme out of this by discretizing the above stochastic
differential equation~(\ref{eq:SPDE-langevin}) (which is formally an
SPDE in the case for infinite dimensional Hilbert spaces). For
simplicity, consider only linear one-step methods to generate a
proposal $y$ from the current $x$, the most general of which is the
stochastic $\theta$-method. When solving the SDE $z'(t)=f(z) +
\sqrt{2}\xi$, this is given by
\begin{equation}
  \label{eq:theta-method}
  z_{n+1} = z_n + \theta \delta f(z_{n+1}) + (1-\theta) \delta f(z_n) + \sqrt{2\delta}\xi_n\,,
\end{equation}
and constitutes a convergent integration scheme for any choice of
$\theta$ (including the explicit and implicit Euler-Maruyama methods
for $\theta=0$ and $\theta=1$, respectively). We now notice that if we
construct $Q(x,dy)$ by discretizing the
S(P)DE~(\ref{eq:SPDE-langevin}) with the
$\theta$-method~(\ref{eq:theta-method}), we obtain
\begin{equation*}
  Q(x,dy) = \mathscr L  \left[\left(1+\theta\delta\right)^{-1}\left(\left(1-(1-\theta)\delta\right) x + \sqrt{2\delta\mathcal C}\xi\right)\right](dy)
\end{equation*}
for $\xi\sim\gamma$, and where $\mathscr L$ denotes the law of its
argument. This yields
\begin{equation}
  \frac{d\nu^\perp}{d\nu}(x,y) = \exp\left(-\frac\delta4\left(2\theta-1\right)\left(\|x\|_{\mathcal C}^2 - \|y\|_{\mathcal C}^2\right)\right)\exp\left(V(y)-V(x)\right)\,,
\end{equation}
so that uniquely the choice $\theta=1/2$ leads to a proposal scheme
without any diverging norms. This is the preconditioned Crank-Nicolson
scheme (pCN)~\cite{cotter-roberts-stuart-etal:2013}, the first
proposal scheme to demonstrate dimension-independent scaling of
acceptance rates.

While this approach treats the case of Metropolis-Hastings of target
measures on Hilbert spaces relative to a Gaussian reference measure,
our situation as presented is considerably more complicated: The
additional modification of the acceptance rate by the projection
scheme introduces similar ratios of diverging norms. Further, as
discussed in section~\ref{sec:cond-expect}, we need to take into
account ratios of curvature terms when considering the restricted
Gaussian measure on the manifold. Both these aspects leads to bad
scaling properties of the algorithm
by~\citet{zappa-holmes-cerfon-goodman:2018}. In special cases, such as
for example on the sphere~\cite{lie-rudolf-sprungk-etal:2023}, these
complications can be overcome, and scalable algorithms can be
constructed, but to our knowledge the problem remains open in
general. For the present work, we restrict ourselves to drawing the
proposal via a dimensionally independent pCN step in the case of a
flat $\mathcal M_z$ without curvature (as in the case of the standard
Brownian bridge). In all other cases, we perform the
finite-dimensional truncations of all steps involved. This yields
acceptable performance for the considered examples, but potentially
implies broken dimensional scaling or broken Markovianity (see for
example~\cite[appendix D]{lie-rudolf-sprungk-etal:2023}).

\subsection{Conditional expectations}
\label{sec:cond-expect}

Unfortunately, it is not our intention to draw from the Wiener measure
$\gamma$, but instead from a conditional Wiener measure $\eta\sim
\gamma|_{F(\eta)=z}$. To make sense of this in terms of Gaussian
measures, we first look at the finite-dimensional case, for which
there is the well-known coarea formula: Given a measurable function
$g:\RR^n \to \RR$ and a continuously differentiable function $F: \RR^n
\to \RR^m$ that foliates $\RR^n$, such that isosets of $F$ are $d=n-m$
dimensional submanifolds of $\RR^n$, we have that
\begin{equation}
  \int_{\RR^n} g(q) \lambda_n(dq) = \int_{\RR^m} \left(\int_{\mathcal M_z} g(q)\, |\det (\nabla F(q)^T \nabla F(q))|^{-1/2} dH_{d}(q)\right)\,\lambda_m(dz)\,,
\end{equation}
where $\lambda_n$ is the $n$-dimensional Lebesgue measure, $H_d$ the
$d$-dimensional Hausdorff measure on $\RR^n$, and $\mathcal M_z = \{ q
\in \RR^n\ |\ F(q)=z\} = F^{-1}(\{z\})$ is the preimage of $z$ under
$F$.

The corresponding version for Wiener space, while considerably more
technical, can be summarized as follows. For $(\mathcal B, \gamma)$
Wiener space with Cameron-Martin space $\mathcal H$, we are given a
map $F:\mathcal B\to \RR^d$. This map is said to be ``non-degenerate''
if the matrix $G_{ij} = \langle D F_i, D F_j\rangle$ is
invertible almost surely, and $G^{-1}$ is smooth. Here $D F$ is
the element of $\mathcal H$ such that $\lim_{\eps\to0}
\eps^{-1}(f(x+\eps h)-f(x)) = \langle D f, h\rangle$. Then,
conditional expectations can be written as
\begin{equation*}
  \EE[g\ |\ F(\eta)=z] = \int_{\mathcal M_z} g(\eta) |\det G(\eta)|^{-1/2} d\gamma_1(\eta)\,,
\end{equation*}
where $\gamma_1$ is the Gaussian measure on the co-dimension 1 fiber
$\mathcal M_z$~\cite{airault-malliavin-ren:2004}. In other words, $|\det
G(\eta)|^{-1/2} d\gamma_1(\eta)$ is the disintegration of
$d\gamma(\eta)$ on $\mathcal M_z$. This means
\begin{equation}
  \gamma(d\eta)|_{F(\eta)=z} = |\det G(\eta)|^{-1/2} d\gamma_1(\eta)\,,
\end{equation}
which is the actual target measure we want to construct our
Metropolis-Hastings algorithm on. Details on the more rigorous
statement of these ideas are given
in~\cite{airault-malliavin-ren:2004, karmanova:2008, malliavin:2015}.

\begin{multicols}{2}
  \footnotesize
  \setlength{\bibsep}{0.0pt}
  
  \bibliographystyle{apsrev4-1}
  \bibliography{bib}
\end{multicols}

\end{document}